# Children's Mental Models of AI Reasoning: Implications for AI Literacy Education

Aayushi Dangol
University of Washington
Seattle, WA, USA
adango@uw.edu

Robert Wolfe
University of Washington
Seattle, WA, USA
rwolfe3@uw.edu

Runhua Zhao
University of Washington
Seattle, WA, USA
runhz@uw.edu

JaeWon Kim
University of Washington
Seattle, WA, USA
jaewonk@uw.edu

Trushaa Ramanan
University of Washington
Seattle, WA, USA
trushaar@uw.edu

Katie Davis
University of Washington
Seattle, WA, USA
kdavis78@uw.edu

Julie A. Kientz
University of Washington
Seattle, WA, USA
jkientz@uw.edu

## Abstract

As artificial intelligence (AI) advances in reasoning capabilities, most recently with the emergence of Large Reasoning Models (LRMs), understanding how children conceptualize AI's reasoning processes becomes critical for fostering AI literacy. While one of the "Five Big Ideas" in AI education highlights reasoning algorithms as central to AI decision-making, less is known about children's mental models in this area. Through a two-phase approach, consisting of a co-design session with 8 children followed by a field study with 106 children (grades 3 - 8), we identified three models of AI reasoning: *Deductive*, *Inductive*, and *Inherent*. Our findings reveal that younger children (grades 3 - 5) often attribute AI's reasoning to inherent intelligence, while older children (grades 6 - 8) recognize AI as a pattern recognizer. We highlight three tensions that surfaced in children's understanding of AI reasoning and conclude with implications for scaffolding AI curricula and designing explainable AI tools.

## 1 Introduction

Few domains of AI research have seen as much recent progress and attention as AI reasoning [59, 78]. In December 2024, OpenAI's o3 large reasoning model set a new state-of-the-art of 87.5% on the Abstraction and Reasoning Corpus (ARC) benchmark [13]. This set of grid-based puzzles is easily solvable by humans, including most children, yet has been historically impossible for even the most advanced general-purpose AI systems. Thus OpenAI's accomplishment led the benchmark's designer to describe the model as "a genuine breakthrough" in AI reasoning [46]. Soon after, the Chinese AI startup DeepSeek created a model known as R1 [19] that is making headlines for its efficient, open-sourced, and high reasoning capabilities.

AI literacy scholarship has long recognized the importance of helping children understand AI reasoning, to the point that *Representation and Reasoning* is one of the AI4K12 Five Big Ideas of AI [44, 84]. However, just as the emergence of large language models prompted a re-evaluation of AI literacy – as children could easily interact with a relatively safe, fluent chatbot to learn with AI [1, 16, 23, 82, 96] – the advent of easily accessible large reasoning models [19, 66], and novel benchmarks designed to measure their reasoning capabilities [36, 43, 60] presents an opportunity to deepen our understanding of children's mental models of AI reasoning and to consider new approaches to teach these concepts. In the present work, we employ the very ARC puzzles used by scholars to evaluate AI reasoning to provide a novel scaffold for understanding children's mental models of AI reasoning. Concretely, we address four primary research questions:

(1) **RQ1**: What kinds of reasoning do children believe AI is capable of, and what do they perceive as the limitations of AI reasoning?
(2) **RQ2**: How can we characterize the mental models of AI reasoning held by children?
(3) **RQ3**: What effects do grade level and prior experience with AI have on children's mental models of AI reasoning?

(4) **RQ4:** How can children's mental models of AI reasoning inform our approaches to AI literacy about models with limited reasoning capabilities, like OpenAI o3 and DeepSeek R1?

To answer these questions, we first conducted a preliminary co-design study with eight children (grades 3 - 8), using a customized interface as a scaffold. The interface included twelve ARC puzzles, which require puzzle-solvers to recognize a visual transformation between grid pairs, and then apply that transformation to a new grid containing a previously unseen patterns (see Figure 1). Considering how AI would perform on puzzles available via this interface allowed children to reflect on AI's reasoning capabilities (*i.e.*, its ability to solve challenging abstract puzzles that demand more than just memorization of data examples) and design novel puzzles that they believed would be challenging for AI. We then conducted a field study with 106 children (grades 3 - 8) that allowed us to more precisely identify children's mental models of AI reasoning and to test the effects of grade level and prior exposure to AI. Drawing on data collected from the co-design study and the field study, our contributions are as follows:

(1) **We find that children expect AI reasoning to be limited in four primary domains**: *Social and Emotional Reasoning*; *Conceptual and Categorical Reasoning*; *Non-Literal Reasoning* (reasoning in settings with linguistic ambiguity, including humor); and *Reasoning with Unfamiliar Representations* of familiar concepts. We provide concrete examples from drawings of novel puzzles produced by children during our co-design study.

(2) **We find that children's mental models of AI reasoning can be characterized as *Inductive, Deductive,* and *Inherent.*** Inductive refers to the view that AI generalizes patterns from data to make predictions; Deductive refers to the view that AI applies predefined rules to reach conclusions based on existing knowledge; and Inherent refers to the view that reasoning capabilities are an intrinsic property of AI, due to its technological nature. We develop these categories based on our co-design study, and we then describe extensive evidence of their presence in the data from our field study.

(3) **We find evidence of a relationship between children's mental models of AI reasoning and their grade level.** We provide statistically significant evidence for the influence of grade level on the type of reasoning children attribute to AI. Specifically, the prevalence of the Inherent mental model becomes less common as grade level increases, while the prevalence of the Inductive mental model increases with grade level. By grade 7, the predominant mental model is Inductive, while the Inherent mental model vanishes entirely, suggesting a shift toward seeing AI as a data-driven pattern recognizer and away from seeing reasoning capabilities as an intrinsic property of AI.

(4) **We offer evidence of three tensions in developing children's literacy about AI reasoning**: the presence of *Overlap and Gaps Between Understanding of Data, Computational, and AI Literacies*; problems with *Generalizing AI Reasoning Across Contexts*; and difficulties in *Balancing AI Literacy with the Pace of Technological Change.* We observed an increasingly challenging environment for AI literacy education, one in which existing approaches to AI literacy foster certain misconceptions about the limitations of AI reasoning in the most recent models. We suggest that, should the rapid pace of change in AI continue, educators will need to equip children with highly flexible understanding of AI, one that is nonetheless grounded in computational and data literacies.

Our work suggests the promise of using simple technological scaffolds like ARC puzzles to further literacy about AI reasoning. It also highlights the opportunity for more integration among approaches to computational literacy, data literacy, and AI literacy, as these branches of technological literacy together inform the mental models through which children will ultimately understand AI.

## 2 Related Work

We first discuss prior research on AI reasoning in K-12 AI literacy and the factors influencing children's understanding and mental models of AI. We then discuss the current state of AI reasoning research and the methods used to evaluate AI reasoning. Given that the term "mental model" has been used in multiple ways across disciplines [35, 64, 70], for the purpose of our study, we draw on Johnson-Laird's framing [45], which conceptualizes mental models as dynamic, situation-specific internal structures that serve as analogs to real or imagined systems. These models are generated on the spot to support reasoning, problem-solving, and explanation, and are shaped by individuals' underlying conceptual structures and prior knowledge [45, 89].

### 2.1 AI Reasoning within K-12 AI Literacy

AI literacy is defined as a set of competencies that enables individuals to "*critically evaluate AI technologies; communicate and collaborate effectively with AI; and use AI as a tool online, at home, and in the workplace*" [52, p. 2]. A key competency in AI literacy is understanding how AI systems *reason*, meaning how AI can "*manipulate representations to derive new information from what is already known*"[84, p. 3]. This understanding is central to forming accurate conceptions of AI [49, 52, 83] and is embedded within the AI4K12 initiative's "Five Big Ideas of AI" [84], particularly the idea of Representation and Reasoning. Teaching approaches in the K-12 level that foreground AI reasoning have emphasized the importance of interrogating AI decision-making processes and engaging children with data and AI models in a hands-on way [25, 47, 58]. For example, Payne [69] worked with young learners to emphasize the importance of training data in machine learning algorithms and helped them explore the potential repercussions of biased datasets on system outputs.

In the past decade, researchers have developed a range of computational and unplugged learning platforms that introduce students to AI's underlying mechanisms [56, 61, 87]. Platforms such as Machine Learning for Kids, Teachable Machine [9], Cognimates [24], and Scratch AI [32] extensions enable children to train AI models, observe predictions, and refine their AI models based on observed outcomes, making abstract AI concepts like classification, AI bias, and model prediction more tangible. Other educational interventions have used an embodied learning approach [14, 37, 50, 54, 63, 81], following Long and Magerko's recommendation [52] that learners can make better sense of an agent's reasoning when they can put themselves "*in the agent's shoes*." For example, Greenwald et al. [37] explored "metacognitive embodiment," where children reflect on their own thinking processes (*e.g.*, emotion recognition) to understand how a facial recognition AI might work. These approaches also align with computational thinking (CT) principles, particularly decomposition and abstraction, as students break down their own reasoning processes to model AI's decision-making [8, 10, 12, 40, 62].

While prior work in AI literacy has made significant strides in helping children understand AI's reasoning processes, the recent development of generative AI models specifically for reasoning introduces new challenges. As described in Section 2.3, large reasoning models have rapidly accelerated progress in AI reasoning. These models still rely on training data, yet they learn strategies for logical reasoning across diverse contexts that were unavailable in previous generations of models. Given that understanding AI reasoning is an important aspect of building overall AI literacy [52, 84], our study starts by seeking to understand children's mental models of AI reasoning and their perceptions of AI's reasoning limitations. By studying the aspects of AI reasoning that children intuitively grasp and where their understanding diverges, we hope to inform the design of novel child-centered AI explanations, educational tools, and interventions to help children develop robust technological literacies.

## 2.2 Factors Influencing Children's Understanding of AI

Children's understanding of AI exists on a spectrum rather than within a binary framework of "right" or "wrong" [57, 67]. Their evolving perspectives reflect both developmental factors and the contexts in which they encounter AI. Prior research shows that age influences how children perceive AI [24, 34, 39], but also that older children do not always exhibit a deeper understanding of AI's mechanisms [88]. While older children (6–10 years old) tend to recognize AI's functional capabilities, they often misattribute intelligence on observable traits such as speed, interactivity, or problem-solving ability [27, 28, 67]. Younger children (3–4 years), by contrast, are more likely to anthropomorphize AI, assigning emotions or intentionality to systems that display responsive behavior [3, 27, 28, 39, 67]. Cultural exposure, socioeconomic status, and parental attitudes further shape these perceptions. Prior research has shown that children in cultures where AI assistants are more integrated into daily life tend to be less skeptical of AI's capabilities [26].

Additionally, the form of AI matters. Flanagan et al. [33] surveyed over 127 children ages 4-11 on their perceptions of AI, using Amazon Alexa and Roomba vacuums as key examples. They found that children view Alexa as having more human-like thoughts and emotions compared to Roomba [33]. Similarly, Dietz et al. [20] explored how adults and children ages 3 through 8 reason about the minds of conversational AI, finding that children do not consistently distinguish between human and AI minds. Moreover, Quander et al. [72] found that children perceive robots with intricate components and dynamic visual cues like flashing lights as more intelligent than robots with simpler designs. Studies of children's understanding of generative AI suggest that children (ages 5–12) perceive it more as a tool for producing content, rather than an entity capable of human emotions [48]. Since children's perceptions of AI may vary by developmental stage and by the type of AI they interact with, our work also assesses these attributes as potential factors contributing to children's mental models of AI reasoning.

## 2.3 The State of AI Reasoning: Large Language Models and Large Reasoning Models

Because our work is ultimately concerned with supporting children's literacy of AI reasoning, we describe here the state of reasoning in AI research, focusing specifically on recent advances in generative AI that have produced the first general-purpose reasoning models by building them on the foundation provided by a large language model. Large Language Models (LLMs) are a form of generative AI trained on a next-word prediction objective, rendering them capable of producing humanlike text [73]. Since the release of ChatGPT in 2022 [65], most LLMs undergo supervised fine-tuning (SFT) that renders them capable of engaging in chat-based conversations with users [85], as well as reinforcement learning from human feedback (RLHF) [68], a process that aligns the model to reflect human social norms (for example, by reducing explicitly biased or toxic output [92, 93]). Much research has demonstrated that LLMs exhibit a limited capacity for reasoning [42]. For example, in a widely used technique called "chain-of-thought reasoning," or COT, a user instructs the model to "think step-by-step" and present the component steps of its reasoning to the user [91]. More complex techniques such as "tree-of-thoughts" use graph-based algorithms to descend LLM outputs generated using COT [99]. COT and the techniques that build on it significantly improve LLM performance on problems that require reasoning.

Large Reasoning Models (LRMs) are LLMs that undergo post-training to generate an internal chain-of-thought (*i.e.*, not part of the output to the user), specifically for the purpose of improving their reasoning capabilities [97]. OpenAI's o1 series of models, released in September 2024, use a reinforcement learning algorithm that rewards a chat-based LLM for using COT to solve difficult reasoning problems [66]. More recently, DeepSeek AI introduced DeepSeek R1 [19], which post-trains the DeepSeek V3 LLM [51] to use COT to solve complex reasoning tasks, without first fine-tuning the model to engage in chat-based dialogue. DeepSeek R1 and OpenAI o1 together set new state of the art on numerous logical, mathematical, and coding tasks [19, 66]. Moreover, analysis of the behavior of these models shows that these models learn to apply reasoning strategies such as re-examining false initial assumptions, spending more time refining an answer before offering it to the user, and breaking down complex tasks into more easily solvable steps [19, 66].

## 2.4 Methods for Evaluating Reasoning in AI

Evaluations of AI reasoning capabilities aim not to assess whether a model has memorized factual information during training, but whether it can apply general reasoning principles to solve complex problems [4]. Unlike benchmarks that *assume* correct answers stem from exposure to relevant training data, reasoning evaluations often use private test sets to prevent models from being trained on specific problem types. These evaluations therefore emphasize a model's ability to generalize reasoning strategies, rather than recall solutions to particular tasks. For example, the GPQA Diamond benchmark maintains a private test set of questions authored by doctoral students that require not only domain-specific knowledge but domain-specific reasoning strategies to answer correctly [75]. Some evaluations detach AI reasoning entirely from factual or domain-specific knowledge. One example is the Abstraction and Reasoning Corpus (ARC) benchmark [13], which we use in this work to probe children's mental models of AI reasoning. As illustrated in Figure 1, ARC presents a model with a series of transformations applied to color-coded grid-based puzzles; the model must infer the rule used in making the transformation and apply it to a new puzzle [13].

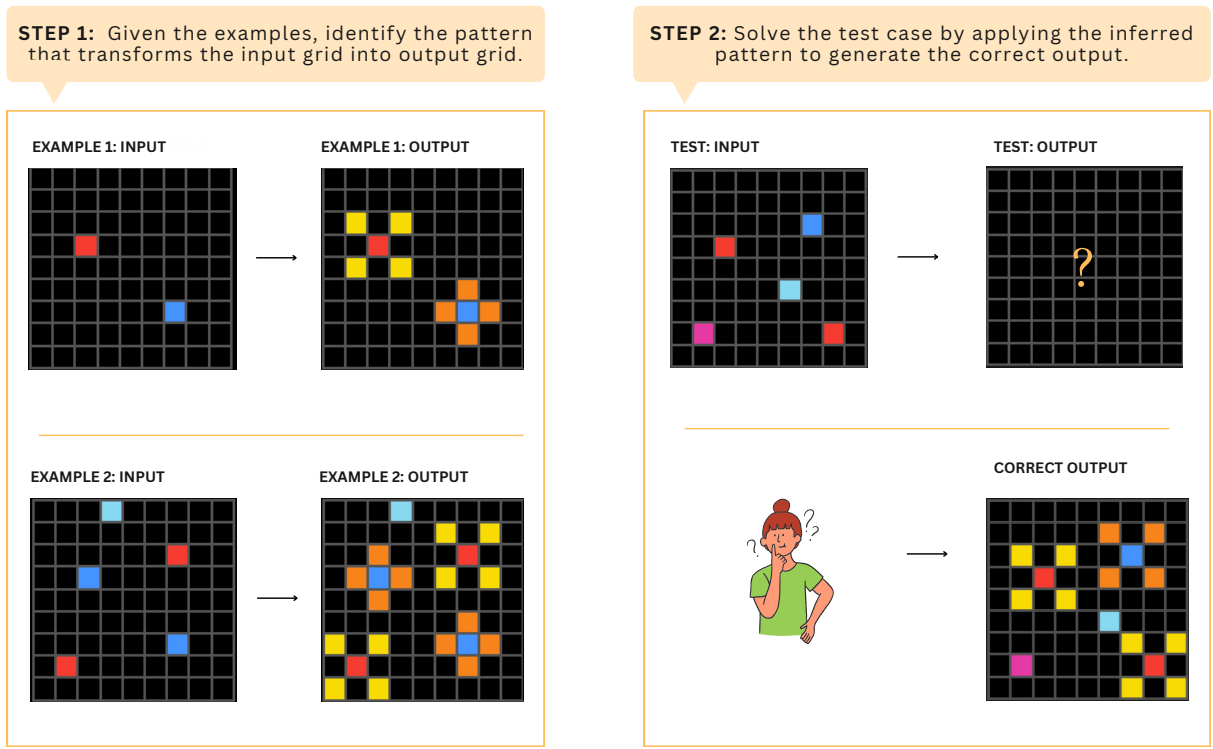

**Figure 1: An ARC puzzle, with two example transformations and a final grid to which the user must apply the rule. In this case, the user must color squares orange if they are directly above or below a dark blue square, or yellow if they are diagonal in any direction from a dark red square while leaving black the squares adjacent to a square of any other color (*e.g., pink or light blue*).**

ARC forces a model to reason about a concrete problem in the absence of relevant factual data on which to draw and without simply applying methods for manipulating text. LRMs, including OpenAI's o1 models, have vastly exceeded the performance of LLMs on these reasoning-specific evaluations [46]. However, even LRMs still commit surprising errors that most humans–including most children–would likely not make [46]. The developers of ARC have promised a more difficult version of the test to more completely evaluate the progress made by LRMs [46], and researchers have developed more complex versions of ARC, such as ConceptARC, which focuses on reasoning with spatial and semantic concepts [60].

## 3 Preliminary Study

Understanding how children conceptualize AI reasoning is essential for designing effective AI literacy interventions. While children engage in reasoning constantly in their everyday lives, whether by testing hypotheses or drawing inferences, they do not use or interpret the term "reasoning" in the same way adults do [79, 80]. To bridge this gap, prior research in cognitive development suggests that using tangible abstractions, such as solving logical puzzles, is a developmentally appropriate way to investigate children's intuitive understanding of reasoning [2, 71]. Thus, as a first step, we conducted a preliminary study to explore whether ARC puzzles could serve as a framing tool to explore how children perceive AI's reasoning processes. By framing reasoning as puzzle-solving, our goal was to examine whether children attribute rule-based logic, probabilistic inference, or other forms of reasoning to AI.

### 3.1 Setting

To test if the ARC Puzzles provided a developmentally appropriate context to surface children's perceptions of AI reasoning, we conducted a preliminary study with an inter-generational co-design group **KidsTeam UW**. We employed Cooperative Inquiry (CI) [29, 30, 102], a participatory method that positions children as equal design partners alongside adult researchers for several reasons. First, CI fosters a democratic environment where children's perspectives are actively requested, valued, and incorporated into the design process [31, 38, 101]. Second, CI has been widely applied in child-computer interaction research to examine how children conceptualize emerging technologies like intelligent interfaces and social robots [17, 61, 94, 95]. Third, children in KidsTeam UW were knowledgeable on multiple participatory design techniques and could, therefore, dive deeply into their design needs [90].

**Table 1: Self-Reported Co-Design Participant Details**

| PID | Gender | Age | Grade | AI Type | AI Use | AI Familiarity |
|-----|--------|-----|-------|---------|--------|----------------|
| P1 | Boy | 8 | 3 | Voice Assistant | Daily | Moderate |
| P2 | Girl | 9 | 5 | None | Never | None |
| P3 | Boy | 7 | 3 | Video Game AIs, Voice Assistant | Daily | Moderate |
| P4 | Boy | 9 | 4 | Video Game AIs, Voice Assistant | Daily | High |
| P5 | Girl | 11 | 5 | Video Game AIs, Voice Assistant | Weekly | Moderate |
| P6 | Boy | 10 | 5 | Chatbot | Weekly | Very High |
| P7 | Girl | 9 | 3 | Video Game AIs, Voice Assistant | Occasionally | High |
| P8 | Girl | 14 | 8 | Chatbot, Video Game AIs, Voice Assistant | Weekly | High |

## 3.2 Participants

Children in KidsTeam UW were recruited through various channels, including mailing lists, posters, and snowball sampling. Once recruited, children participate throughout the school year, attending one or both of the weekly sessions offered by the group. Table 1 provides demographic information for the eight children who participated in our preliminary study. Children reported varying levels of AI familiarity and use. Some participants regularly interacted with AI through voice assistants (*e.g.*, Alexa, Siri) and video game AIs, while others engaged with chatbots or had no direct AI experience. We obtained parental consent and child assent for all participants, and our university's Institutional Review Board (IRB) reviewed and approved all research related activities with KidsTeam UW.

## 3.3 Materials

To facilitate children's exploration of ARC puzzles, we developed a web-based application accessible through any modern browser [18]. The application featured 12 curated puzzles from the ARC dataset [13], organized into four levels of increasing difficulty. Figure 2 depicts participants solving ARC puzzles using the web-based interface.

While many logical puzzles could serve as scaffolds, we chose ARC puzzles for several reasons. First, ARC puzzles are designed to be content-agnostic, meaning they do not require domain-specific knowledge such as mathematical formulas or linguistic proficiency. This ensures that children from diverse backgrounds can participate meaningfully. Second, ARC puzzles allow for a clear separation between learning and inference. Because each puzzle presents new transformations that must be discovered, this property helps investigate if children perceive AI's reasoning as relying on learned patterns, inferential, or a combination of both. Third, ARC puzzles are highly interpretable. Unlike many reasoning tasks where AI solutions may be opaque or require complex explanations [36, 75], ARC puzzles provide a clear visual representation of both the problem and the solution. This makes them particularly useful for eliciting children's explanations of reasoning, as they can point to concrete transformations rather than abstract verbal descriptions. Finally, ARC puzzles are widely recognized as a benchmark for reasoning in AI research. Since the puzzles are used in evaluations of both LLMs and LRMs, they provide a standardized measure against which children's mental models of AI reasoning can be compared to AI's actual reasoning process.

## 3.4 Procedure

Our 1.5-hour co-design session was structured to balance relationship-building, discussion, and hands-on design activities. We began with **Snack Time** (15 minutes), where all eight children and facilitators sat together in an informal, shared space. This time was intended to build rapport and trust, allowing children to settle in and socialize with each other and the adult facilitators. Next, during **Circle Time** (15 minutes), an adult facilitator introduced ARC puzzles, using a large display and verbal walkthrough to demonstrate how to solve a puzzle. This whole-group introduction ensured all children had a shared understanding of the activity before transitioning into smaller groups.

Once the children understood the rules for solving ARC puzzles, the children were split into three small groups – two groups of three children and one group of two – each supported by two facilitators. As children solved the 12 ARC puzzles using the web-interface (as discussed in Section 3.3), they were encouraged to verbalize their thought processes, explaining the patterns they noticed, the rules they inferred, and how they applied them in trying to solve the puzzles. Facilitators asked questions such as, *"Do you think AI could solve these puzzles?"* and *"How would AI solve it?"* to understand their perspectives on AI reasoning. We then transitioned into a participatory design activity called *Likes, Dislikes, and Design Ideas* [38], wherein adult facilitators captured children's responses about what they liked about the puzzles and what they disliked, and invited them to share any ideas for modifying the puzzle-solving experience.

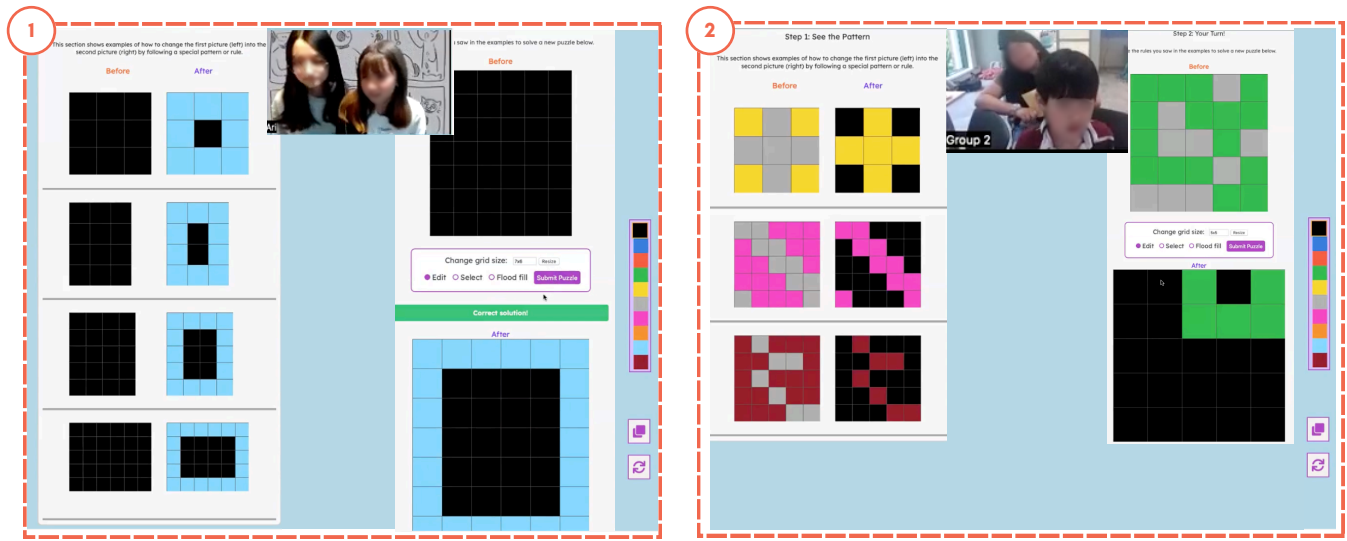

**Figure 2: Participants engaging with the web-based ARC puzzle interface. Screenshots show participants solving (1) a Level 1 puzzle and (2) a Level 4 puzzle.**

Following this, we introduced a second design prompt: *"What kind of puzzle would be easy for you but hard for AI?"* This prompt served as a scaffold for children to articulate their understanding of both their own problem-solving processes and AI's reasoning limitations. Children continued working in their small groups or individually if they preferred, engaging in an open-ended design process. They could sketch puzzle ideas or describe them in writing, with facilitators supporting them in articulating their concepts. Finally, we reconvened as a whole group for **Big Ideas** (15 minutes). Children took turns presenting their puzzle designs, explaining their solutions, and discussing why they believed their puzzles would challenge AI. This group reflection allowed children to share insights and build on each other's ideas about human versus AI reasoning.

### 3.5 Data Collection

During the co-design session, our team utilized built-in webcams on desktop computers to record video and screen interactions using the Zoom video conferencing software. Three cameras recorded three separate groups of eight children, capturing a total of 93 minutes of video. To transcribe our video data, our research team created analytical memos summarizing key interactions and discussions [6, 76]. The fourth author watched the recordings and documented notable events at 5-minute intervals, capturing children's interactions with the ARC Puzzles and children's dialogues about how their problem-solving approach would compare to that of AI. After the fourth author finished writing memos, the first and third authors independently reviewed the same videos and the fourth author's memos to verify the accuracy of the initial observations and to add additional notes and insights. This dual-review process supported the reliability of the data analysis process and allowed us to capture more than one perspective on the content of the videos.

### 3.6 Data Analysis

Following memo creation, we followed a thematic analysis approach [7] to interpret the data. The first and second authors began by independently reviewing the analytical memos, suggesting initial codes such as "AI Reasoning Challenges" and "AI Pattern Recognition." They then held a series of three meetings to reconcile codes, collaboratively review participant quotes, examine counter-examples, and refine the boundaries and definitions of each code. For example, the code "AI Reasoning Challenges" was initially broad, encompassing various difficulties children believed AI would face. However, after analyzing participant responses, we refined this category into four specific reasoning challenges: "Non-Literal Reasoning", "Unfamiliar Representations", "Emotional Reasoning", and "Conceptual Reasoning." During this process, overlapping codes were also systematically merged and organized into broader categories. For example, the codes "AI Pattern Recognition" and "Learning from Data" were merged into a single category, "Inductive Reasoning." After refining the codebook three times, our final codebook included six codes and 14 subcodes. The first author then applied the final codebook to the full dataset, and the second author independently reviewed the coded data to ensure comprehensive analysis. Once all the data were coded, the first and second authors met to discuss and resolve any coding disagreements. We then organized the codes into overarching themes, and the first author revisited the entire dataset to extract representative quotes for each theme.

## 4 Co-Design Findings

We present findings from our preliminary study, including children's engagement with ARC puzzles, their perceived limitations of AI reasoning, early evidence of three mental models of AI reasoning, and key challenges in supporting AI reasoning literacy among children.

## 4.1 Children's Engagement & Feedback on ARC Puzzles

Our participants exhibited high levels of engagement when solving the ARC Puzzles. Across all groups, children actively discussed puzzle transformation rules and tested different approaches to solving the puzzles. During the *Likes, Dislikes, and Design Ideas* activity, children expressed appreciation for the reasoning challenge posed by the puzzles. For example, P1 (boy, age 8, grade 3) stated, "*I like how you had to look at the clues to find the answer,*" while P7 (girl, age 9, grade 3) said that "*pattern recognition is easy enough for a second-grade kid except the last puzzle.*" Children also identified usability challenges. While we set up the puzzles to be solved on desktop computers, children found repeatedly clicking the grid cumbersome and preferred touchscreens. For example, P3 (boy, age 7, grade 3) said, "*Put in a touchscreen, clicking is annoying.*" Other participants had difficulty identifying the configuration buttons for navigating between puzzle levels, and they expressed interest in a "cloning" feature that would allow them to duplicate and then edit the input grid. We addressed all of these concerns to improve interaction for our subsequent field study. We also found that organizing the puzzles into four levels of increasing difficulty—each containing a set of three puzzles—helped maintain engagement, as children and expressed a sense of accomplishment upon solving more challenging puzzles.

## 4.2 Children's Perceived Limitations of AI Reasoning

### 4.2.1 Reasoning with Unfamiliar Representations.

Children believed that AI would struggle with reasoning about puzzles that employed inputs presented with atypical representations or in unfamiliar formats. For example, P3 (boy, age 7, grade 3) designed a puzzle that required AI to create a circle by composing it out of only square pixels on a square grid (see Figure 3). When asked to explain why AI would struggle with this, P3 explained, "*it will want to make it perfectly round, but it's impossible for AI to make it circle while the two materials are both in square shape.*" P3's puzzle requires an understanding that a circle can be represented approximately using squares. This requires an ability to reinterpret the concept of a circle in a flexible way, regardless of the apparently unsuitable input. A rigid approach, which P3 attributed to AI, would conclude that a perfect circle is impossible and fail to recognize an approximate solution.

Similarly, P2 (girl, age 9, grade 5) designed a puzzle that required a visual - rather than purely numerical - approach to solving otherwise straightforward mathematical problems (see Figure 4). The puzzle represented numbers using grids of color-filled squares. To find the correct solution, AI had to manipulate the quantity of filled squares in accordance with several mathematical operations, such as increasing or decreasing the number of squares to perform addition or subtraction respectively. P2 explained that this puzzle was difficult because AI needed to carefully track and adjust the number of filled squares, a process she believed would be "*hard for AI.*" To further challenge AI, P2 imposed a strict rule stating that "AI could only use the grid to solve the problems, and if a mistake was made, it had to start over." P2's rule reflects an interesting awareness of AI's ability to correct initial mistakes during its reasoning process,

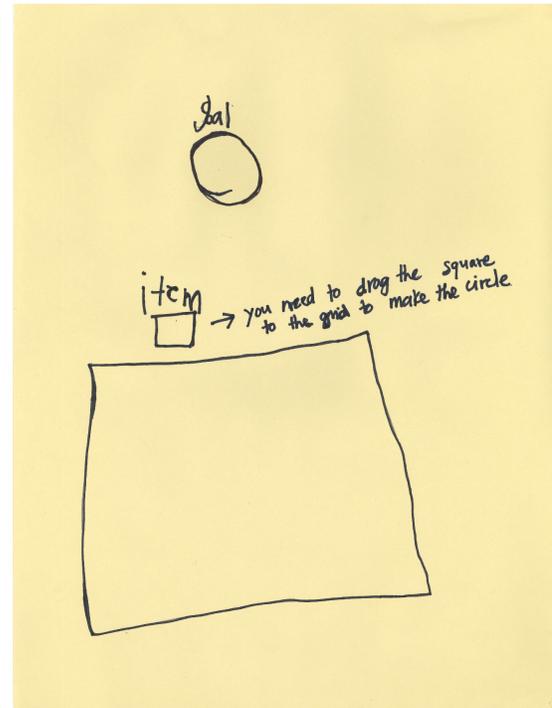

**Figure 3: Puzzle designed by P3 (boy, grade 3) that challenges AI to create a circle using only square pixels on a square grid.**

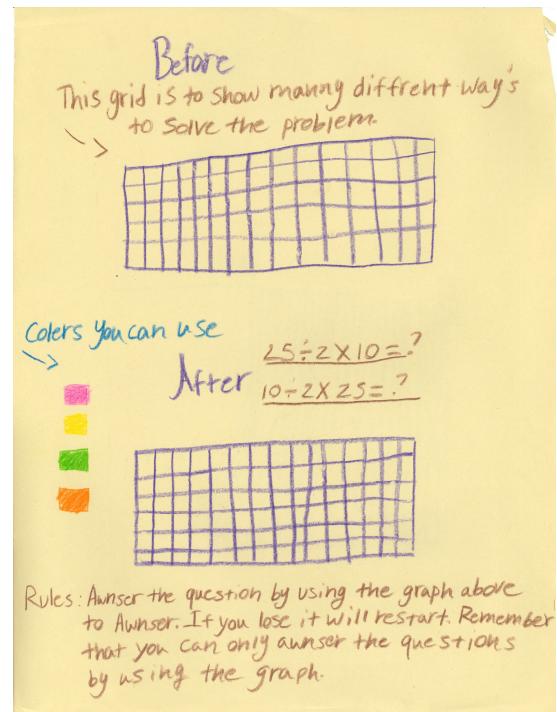

**Figure 4: Puzzle designed by P2 (girl, grade 5) that requires AI to solve numerical problems using visual grids.**

which is a hallmark of large reasoning models like DeepSeek and o1.

*4.2.2 Social and Emotional Reasoning.* Children believed that AI would struggle with puzzles that required an understanding of human social dynamics and emotions. For example, P7 (girl, age 9, grade 3) designed a puzzle that required an understanding of relationships (see Figure 5).

Her puzzle posed the question, "*How do you start a relationship?*" and provided four possible answers: "*a) fight, b) get to know each other, c) walk away, and d) nothing.*" P7 later explained that "*to solve the puzzle, you have to know relationships. AI doesn't know about relationships.*" This perspective highlights the belief that AI's lack of human social experiences will make reasoning about such situations challenging.

*4.2.3 Non-Literal Reasoning.* Several child participants believed that AI would struggle with non-literal reasoning, which refers to the ability to understand and interpret meanings beyond the explicit, surface-level information presented in a problem. P1 (boy, age 8, grade 3) designed a verbal riddle to test AI's reasoning: "*You are in a box, and the only item you have is a famous baseball bat. How do you get out?*" P1 then revealed the answer: "*You try to hit the ball with the bat three times,*" referring to striking out in baseball, where a batter is out after three missed swings. P1 believed this was an "*impossible puzzle*" for AI because it wouldn't "*get the joke.*" The humor in his riddle relies on the double meaning of "getting out," a nuance he believed AI would struggle with because non-literal reasoning is necessary to understand jokes and interpret double meanings.

*4.2.4 Categorical and Conceptual Reasoning.* Finally, children also believed that AI would struggle to reason about abstract categories or concepts. P8 (girl, age 14, grade 8) designed a "*spot the difference*" puzzle (see Figure 6), as she thought that AI might struggle to identify odd-one-out images because it doesn't "*know much about concepts humans talk about,*" which would require "*living experience.*" She initially proposed a puzzle featuring four characters: "*a ghost, a skeleton, a unicorn, and a witch.*" When an adult facilitator guessed that the unicorn was the odd one out because it wasn't scary, P8 reconsidered the puzzle's difficulty, believing it to be too easy. She refined the design by using only "*fictional characters to make the distinction even harder.*" Ultimately, P8 settled on a puzzle "*of a zombie, a witch, a unicorn, and a ghost.*" She said that "*AI will struggle because solving this puzzle needs human experience that is outside of logical thinking.*"

## 4.3 Children's Mental Models of AI Reasoning

In addition to surfacing children's perceived limitations of AI reasoning capabilities, our co-design study suggested the existence of three models of AI reasoning among children: 1) Inductive reasoning, where AI generalizes patterns from data to make predictions, 2) Deductive reasoning, where AI applies predefined rules to reach conclusions based on existing knowledge and 3) Inherent reasoning, where AI is perceived as naturally capable due to its technological nature.

Several children described AI as capable of Inductive reasoning, believing that AI learns to apply general rules based on examples. P8

(girl, age 14, grade 8), for example, said, "*If AI is given the examples beforehand, it could probably solve harder puzzles.*" P6 (boy, age 10, grade 5) echoed this sentiment, noting that "*It [AI] will learn from his and other people's answers,*" suggesting that P6 viewed even the KidsTeam UW group's responses to the puzzles as potential AI training data. Similarly, some children conceptualized AI as engaging in Deductive reasoning, emphasizing that "*AI is coded by humans* (P2, girl, age 9, grade 5)" or "*I think that AI can solve puzzles without having problems because they are programmed to do so* (P4, boy, age 9, grade 4)," indicating a belief that AI follows rules programmed by humans. Others exhibited an Inherent reasoning perspective, perceiving AI as all-knowing. For example, P7 (girl, age 9, grade 3) stated that "*AI can [solve puzzles] because it knows everything.*" When the facilitator asked P7 to elaborate more, she justified her belief by emphasizing AI's speed, responding,"*[An adult] was stumped on the last one; an AI could have gotten that one faster.*" Based on these initial observations, we deductively coded the data collected for our Field Study (described in Section 5) as reflective of an Inherent, Inductive, or Deductive mental model of AI reasoning.

## 4.4 Challenges of Developing AI Reasoning Literacy

*4.4.1 Addressing a Wide Variety of AI.* We found that children were aware of many different forms of AI and discussed them in our co-design session. Some students discussed chatbots, while others spoke about voice assistants. In group discussions, this meant that children often considered AI reasoning primarily by thinking about their prior experience with certain specific forms of AI. Additionally, some children also saw AI as existing on a spectrum ranging from simple to highly advanced. For example, P8 (girl, age 14, grade 8) stated, "*It depends on the AI. If we're talking about really intelligent AI, then I agree, but an undeveloped AI might not be able to.*" This observation during the preliminary study led us to hypothesize that prior experience with AI might influence children's mental models of AI reasoning, which we further explored in our field study.

*4.4.2 Negotiating Computational and Data Literacies.* Second, we observed that students brought different underlying computational literacies to bear during discussions. For example, many children approached AI reasoning through the lens of data literacy: they believed that if AI had seen examples of the puzzles before, it would be able to solve them. On the other hand, some children approached AI reasoning through the lens of computing literacy. These children tended to believe that AI would be capable of reasoning if a human had programmed it to. Though both are important and relevant perspectives, they encourage very different understandings of AI. Moreover, neither perspective actually captures the way modern reasoning models are trained: by defining a "reward function" that forces models to engage in deductive reasoning over problems from highly varied logical and mathematical domains.

*4.4.3 Staying Current With a Rapidly Changing Technology.* Finally, we found that many of the children assumed that AI could not perform tasks that, in many cases, it now can. For example, when children were gathered for group discussion, P7 (girl, age 9, grade 5) said, "*I hope it doesn't Google answers to puzzles*" to which P1 (boy, grade 3) replied "*AI don't Google,*" while P2 (girl, age 9, grade 5) remarked, "*Impossible because it can't Google.*" Though the children were right in that LLMs and LRMs would not typically use Google or the internet to solve block puzzles, such models now have the capacity to search the web and retrieve information. This points to the difficulty of updating AI education curricula in light of the fast-moving nature of AI.

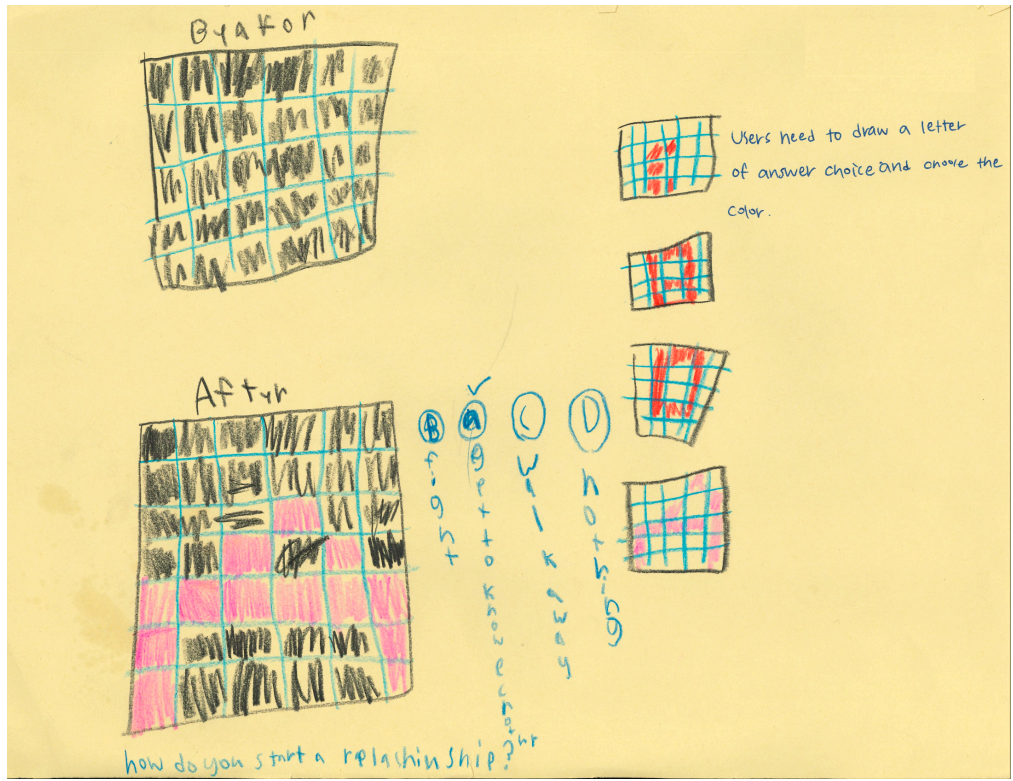

**Figure 5: A puzzle designed by P7 (girl, grade 3) that requires social and emotional reasoning. The puzzle presents the question, "How do you start a relationship?" with four possible answers: a) fight, b) get to know each other, c) walk away, and d) nothing. Players must draw a letter corresponding to their chosen answer in the grid.**

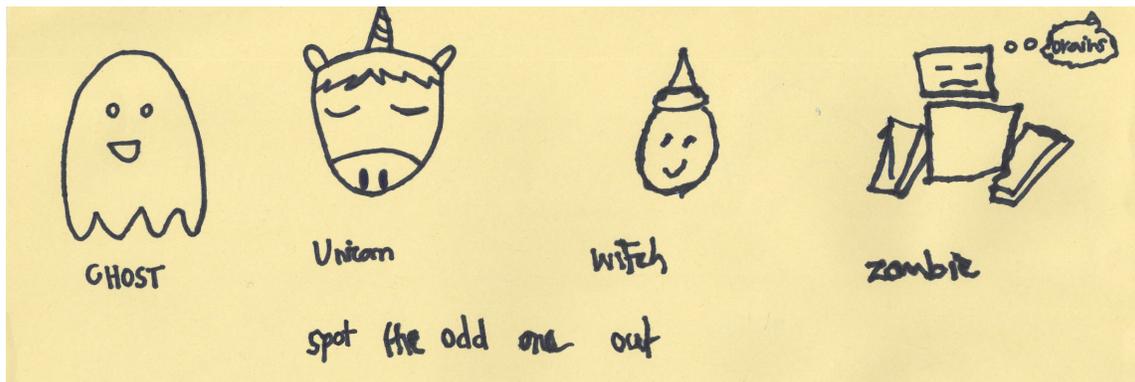

**Figure 6: A "spot the difference" puzzle designed by P8 (girl, grade 8) that requires categorical and conceptual reasoning. The puzzle features four fictional characters: a ghost, a unicorn, a witch, and a zombie, asking players to "spot the odd one out."**

## 5 Field Study

Children's high engagement with ARC puzzles in our preliminary study supported their use as a developmentally appropriate and engaging tool for surfacing children's perceptions of AI reasoning. Their suggestions from the co-design session also led us to refine the puzzle interface to support smoother interaction. Building on these insights, we conducted a field study with 106 children (grades 3–8).

This larger-scale study helped to extend our preliminary findings by capturing the perspectives of a broader and more diverse group of children than we could have with a small-*N* lab study [100]. It also allowed us to examine from a quantitative perspective whether factors such as grade level and prior interactions with AI influence children's mental models of AI reasoning.

## 5.1 Setting

The field study took place at **UW Discovery Days**, an annual outreach event at our university that invites members of the local community to campus. Attracting approximately 10,000 children from the local community as well as their teachers, parents, and guardians, the event intends to foster community engagement with STEM fields specifically. The event is targeted toward grades 4 through 8, though children outside that grade range may also attend as siblings. A designated space within the venue served as the study site, where our study team set up a booth called "Puzzleland." The study setup included two stations where children engaged with ARC puzzles, plus an additional station for obtaining assent and background information from participants and their chaperones. The setup allowed for naturalistic engagement, as children and families voluntarily approached our booth, creating an informal and exploratory environment similar to exhibits in museums or science centers.

## 5.2 Participants

We had 106 child participants engage with our study activity. The study was reviewed by our university's IRB and determined to be exempt since the activities were educational in nature, and thus consent was not required. Nonetheless, we still obtained written assent from the children and verbal consent from their chaperones (who may have been teachers, parents, or other adults entrusted with their care) prior to participation. Table 2 summarizes the demographics of the participants, whose grade levels ranged from 3rd grade through 8th grade. In the U.S., most 3rd graders are 8–9 years old, and most 8th graders are 13–14 years old. The gender distribution was as follows: 57.5% girls, 40.6% boys and 1.9% preferred not to say. Participants reported interacting with various AI technologies, with Voice Assistants (52.8%) and Video Game AIs (45.3%) being the most commonly used.

## 5.3 Procedure

Potential participants were invited to take part in the study as they approached the Puzzleland booth. Interested children and their chaperones were guided to a designated table, where a researcher explained the study using simple, accessible language. The researcher ensured that each child understood their participation was voluntary and that they could stop at any time. Children who agreed to participate signed a written assent form and completed a short background survey (with help from their chaperones, if needed). We also obtained verbal consent from each child's accompanying adult, who was informed about the study's purpose, duration, and procedures. Chaperones were welcome to stay and observe or wait nearby while the child participated. Child participants solved ARC puzzles individually using the interface described in Section 3.3. The interface was available on touchscreen tablets

**Table 2: Reported Survey Participant Demographics**

| Social Category | Participant Demographics (*n*=106) |
|---|---|
| Gender | Girl (57.5%), Boy (40.6%), Prefer Not to Say (1.9%) |
| Grade | 3 (1.9%), 4 (27.4%), 5 (28.3%), 6 (22.6%), 7 (12.3%), 8 (7.5%) |
| AI Use* | Voice Assistants (52.8%), Video Game AIs (45.3%), Personalized Recommendations (38.7%), Chatbots (32.1%), Not Sure (14.2%), No AI Use (13.2%) |
| AI Familiarity | None (4.7%), Low (23.6%), Moderate (30.1%), High (35.8%), Very High (4.7%) |

*Note that many participants could report using more than one type of AI.*

(iPads) and laptops provided on-site, both featuring the same set of 12 puzzles. Participants could choose the device they found most comfortable. While all children were asked to solve at least two puzzles, they were free to explore and attempt as many additional puzzles as they liked, in any order.

Although all children in our preliminary study were able to complete the puzzles, it was still important to ensure that every child left the Puzzleland booth with a sense of achievement. To support this goal, if a child was not making progress on a puzzle (*e.g.*, if we observed them repeating the same reasoning strategy unsuccessfully), they were provided with progressively more specific hints by a researcher at the booth. After completing the puzzles, each child was given a paper worksheet with a prompt asking whether they believed AI could solve the ARC puzzles and, if so, explaining how it would. At the conclusion of the study, each child was thanked for their participation and could choose between a sticker pack or a university-branded pen.

## 5.4 Data Collection

We collected two primary forms of data: background surveys and post-puzzle reflection. The survey captured information such as grade level and gender, as well as prior experience with AI technologies. Participants could select from a list of AI types they had used (e.g., voice assistants, chatbots) and rate their familiarity with AI on a 5-point scale. After completing the ARC puzzles, each participant received a reflection prompt asking whether they thought AI could solve the ARC puzzles, and if so, how. These open-ended responses provided insight into how children interpreted the reasoning involved in the task and their assumptions about AI capabilities.

## 5.5 Data Analysis

Our analysis focused on children's written reflections about whether and how they believed AI could solve the ARC puzzles. We began with the codebook developed during our preliminary study as a foundation. This codebook included three primary reasoning types: Inductive Reasoning, Deductive Reasoning, and Inherent Reasoning (see Table 3 for definitions and examples). To remain open to new insights beyond these predefined categories, we also allowed for emergent codes during the coding process. However, no additional categories of reasoning were consistently observed across participants.

The first and second authors independently coded the entire dataset. The inter-rater agreement was assessed using Cronbach's alpha, yielding a value of 0.84, indicating high reliability. They then met over two meetings to discuss and resolve any coding disagreements [55]. In light of the large dataset (over 100 children), we also explored the possibility of finer-grained distinctions within each reasoning type. This analysis surfaced subtle variations—for instance, some children drew on informal, personal encounters with AI, while others referenced their formalized understanding shaped by prior computing education. While these subtypes did not warrant new top-level codes, they inform our discussion of how children operationalize different forms of reasoning. Following our qualitative analysis, we used the chi-square test to investigate potential relationships between children's grade level, their previous experience with AI, and their mental models of AI reasoning (*i.e.*, Inductive, Deductive, and Inherent Reasoning). Results were evaluated for statistical significance at p < .05.

## 6 Field Study Findings

In this section, we discuss the findings of our field study. We first report the results of our qualitative analysis, wherein we identified three primary mental models of AI reasoning held by participants. We then report the findings of our statistical analysis of whether a relationship exists between children's mental models and either their grade level or their prior use of AI. Finally, we discuss our findings with regard to how children's mental models inform their perception of the limitations of AI.

### 6.1 Children's Mental Models of AI Reasoning

Our thematic analysis provided evidence for three primary mental models of AI reasoning among our field study participants. We referred to these mental models as *Inherent Reasoning*, *Inductive Reasoning*, and *Deductive Reasoning*, corresponding to the reasoning capabilities perceived by children to be possessed by AI.

*6.1.1 Inherent Reasoning.* 34 of our 106 child participants (32.0%) conceptualized AI reasoning as an intrinsic ability, independent of its programming, exposure to data, or capacity for pattern recognition to solve puzzles. Their responses indicated an equivalence between reasoning and AI - because AI is artificial *intelligence*, it must have the capacity to reason. Examples of the *Inherent Reasoning* mental model manifested in several ways in children's responses. In the first, children viewed reasoning as something AI simply does without specifying a clear mechanism for how it works. For example, P75 (girl, grade 5) stated, "*AI can solve them [the puzzles] because they are really smart,*" while P90 (girl, grade 5) similarly said that AI can solve the puzzles "*cause they are robots and are very smart.*"

The Inherent Reasoning mental model also manifested in abductive inferences made by children about AI: based on observations of AI's output, the simplest explanation is that the model must possess reasoning abilities. Children made three kinds of abductive inferences, the first of which identified the speed and efficiency of

AI responses as evidence of reasoning ability. For example, P119 said, "*AI can solve them because it is quicker to find the patterns than humans*" (boy, grade 6), while P20 said, simply, "*because it solves faster*" (girl, grade 4). These and similar responses suggest that for some children, reasoning is tied to processing speed–they see the ability to respond to a problem quickly as evidence of intelligence. Additionally, some children associated AI's generative capabilities with its ability to reason. P4 (girl, grade 5) explained, "*AI can figure things out because ChatGPT can generate an answer,*" suggesting that the participants perceives the ability to create human-legible responses as a form of reasoning. Some children took this idea further, speculating that "*AI can solve the puzzles because it probably made the puzzles so it can solve them*" (P24, girl, grade 4), implying that AI's reasoning stems from an assumed built-in knowledge of its own creations. Finally, some children drew from past experiences of observing AI handling complex tasks to justify their belief in its reasoning. For example, P78 (boy, grade 4) stated, "*It can do tenth-grade calculus, and chemical equations, and I have seen it before,*" while P39 (boy, grade 4) added, "*Puzzles are very hard, and despite that, I've seen AI [solve them].*" Rather than considering the kind of reasoning performed by AI, the observations above focus on establishing *that* AI can reason, and that this reasoning ability is intrinsic to AI.

*6.1.2 Inductive Reasoning.* A total of 38 child participants (35.9%) conceptualized AI's reasoning as inductive. These participants foregrounded the role of data as the source of AI's capabilities. They viewed AI as reasoning by recognizing patterns, making predictions based on what it learns from data, and improving its representations over time when presented with new data. P80 (boy, grade 5) stated, "*AI observes and trains patterns,*" while P132 (girl, grade 7) explained, "*AI is programmed to recognize patterns in data.*" Similarly, P120 (boy, grade 7) said, "*With recent advancements, AI has had massive leaps into interpreting data.*" Some children also identified AI's ability to learn new rules on the basis of new data. P133 (boy, grade 8) noted, "*They [AI] can learn over time by trying over and over again,*" and P92 (boy, grade 6) emphasized, "*AI gets better over time and it's [sic] program can adapt.*"

Older children in particular discussed the importance of *training* data, recognizing that AI does not simply follow pre-programmed rules but instead infers patterns from data. They focused on the role of large datasets in AI's training process. For example, P84 (boy, grade 6) stated, "*if it gets enough data to train, then it can learn,*" while P114 (girl, grade 6) said, "*they need to be trained and have a really big data set.*" These participants understood AI's reasoning as formed during a training phase, where AI learns patterns and solutions from vast amounts of pre-existing data. This perspective indicates an awareness of some machine learning principles, where AI is perceived as dynamic rather than static, capable of refining its reasoning based on accumulated experience.

*6.1.3 Deductive Reasoning.* Of the 106 child participants, 34 (32.0%) conceptualized AI reasoning as deductive. These participants perceived AI as applying specific rules given a set of premises. Some children with this view of AI observed that AI's responses were dependent on the input (*i.e.*, the prompt) provided by the user, and that, given the right input, AI would logically arrive at the correct output. For example, P97 said AI could solve the block puzzles

**Table 3: Definition of Codes and Example coding**

| Code | Definition | Example Coding |
|---|---|---|
| **Inductive Reasoning** | AI generalizes patterns from data to make predictions. | *"AI can because they are programmed to recognize patterns."* (P132, girl, grade 7) |
| **Deductive Reasoning** | AI applies predefined rules to reach conclusions based on existing knowledge. | *"Because AI is programmed with knowledge."* (P72, girl, grade 5) |
| **Inherent Reasoning** | AI is perceived as naturally capable due to its technological nature. | *"Because they are robots."* (P17, girl, grade 4) |

*"Because you have to give it instructions and if it is given the right information about how to solve them it can"* (P97, girl, grade 6). Participants like P97 place the onus on the user to provide AI with the right set of premises, rather than on AI to learn the right logical rules during its training process.

However, many participants who perceived AI reasoning as deductive curiously located AI's reasoning capabilities outside of the model itself. For example, some children viewed deductive reasoning capabilities as an extension of AI programmers' abilities. For example, P44 (girl, grade 4) said, *"If the people who program it can, AI can do it too,"* while P13 (girl, grade 3) said, *"AI can solve the puzzles because they are coded to be able to."* Other children framed AI's problem-solving as being dependent on the ability to retrieve information from the internet, rather than only using internal reasoning abilities. P79 (boy, grade 5) said *"Alone, it cannot use logical reasoning, but it can connect to similar puzzles online, it can solve it,"* while P72 (grade 6) said *"AI can easily solve the block puzzles because it has the entire web at its access."* Though these participants stop short of saying that AI simply retrieves information from the internet, they nonetheless link deductive reasoning to external data.

## 6.2 Effects of Grade Level and Prior AI Exposure on Mental Models of AI Reasoning

*6.2.1 Does Grade Level Influence Children's Mental Models of AI Reasoning?* We used a chi-square test to evaluate the relationship between children's grade level and their mental model of AI reasoning (*i.e.,* deductive, inductive, or inherent). We obtained a statistically significant result of $\chi^2(10) = 32.00$, $p < .001$, with a moderate effect size of Cramer's $V = 0.39$, providing evidence for the influence of grade level on the type of reasoning children attribute to AI. As illustrated in Figure 7, the proportion of children whose responses indicate an Inherent mental model declines steadily across grade levels, and the model disappears entirely from our data after grade 6. Conversely, the proportion of children whose responses indicate an Inductive reasoning model increases progressively, becoming the predominant perspective among our respondents by grade 7. This suggests a developmental shift: younger children (grades 3 - 5) tend to see AI reasoning as intrinsic to the technology, while older children (grades 6 - 8) begin to perceive AI as a system that learns from patterns and data. We also observe some evidence for an increase in the proportion of children whose responses indicate a Deductive reasoning model from grade 6 to grade 7. While the increase is not durable in grade 8, we collected only 8 responses from eighth graders, significantly less than for other grades. Moreover, while we see some evidence for a Deductive reasoning model among younger children, we note that some of these perspectives reflect less accurate mental models of AI, such as the perception that AI reasoning is the result of rule-based programming by humans.

*6.2.2 Does AI Type Influence Children's Mental Models of AI Reasoning?* We employed chi-square tests to evaluate whether the type of AI used by our participants (voice assistants, chatbots, video game AIs, and/or personalized recommendation systems) had any effect on their mental model of AI reasoning. However, none of our chi-square tests were statistically significant at a significance level of $p < .05$. Of the types of AI considered, chatbots ($\chi^2(10) = 5.16$, $p = .076$) and videogame AI ($\chi^2(10) = 5.29$, $p = .071$) exhibited associations at a level that might be considered trends. Moreover, given that some children reported using multiple forms of AI, we further investigated whether there was a statistically detectable relationship between mental models of reasoning and the use of only one AI type (narrow AI users); the use of multiple AI types (broad AI users); and the use of no AI type (no AI users). However, the results of these chi-square tests were not statistically significant at the level of $p < .05$. Thus, we also found no evidence of a relationship between children's breadth of AI use and their mental models of AI reasoning. We found this surprising, primarily because we expected that prior exposure to chatbots, in particular, might yield evidence of such a relationship, especially given that chatbots, in some cases, lay out their reasoning in a step-by-step format for the user.

## 7 Discussion

Findings from our study identify three mental models of AI reasoning: Inductive, Deductive, and Inherent. While younger children (grades 3 - 5) often relied on observable traits such as speed and efficiency to describe AI's reasoning as inherent, older children (grades 6 - 8) demonstrated an emerging understanding of AI concepts such as "pattern recognition" and "training data." However, misconceptions persisted across all grade levels, showing that children struggle with three tensions around AI reasoning that educators and researchers can take note of:

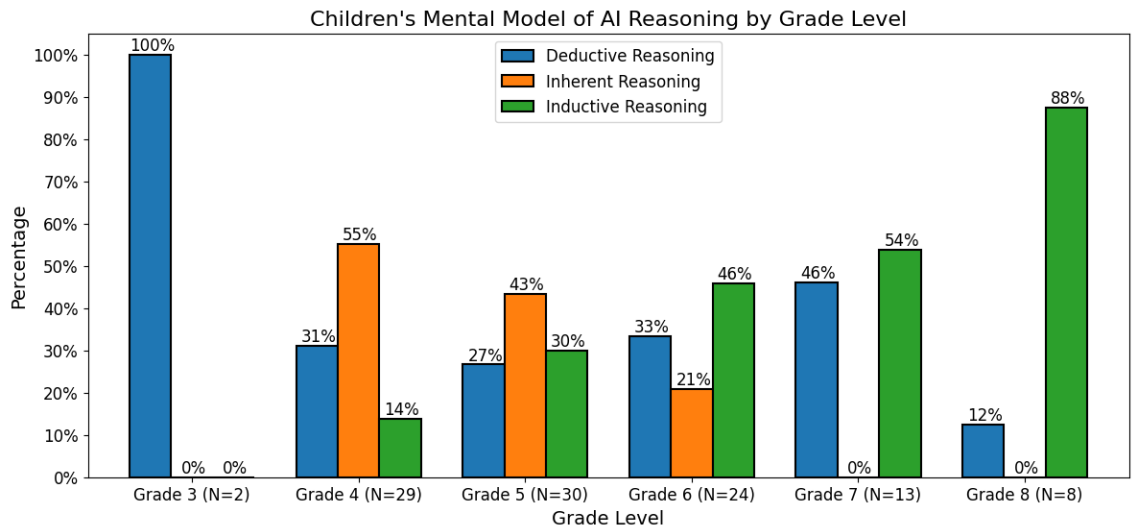

**Figure 7: We found that the proportion of children whose responses indicated an Inherent reasoning mental model declined as grade level increased, while the proportion whose responses indicated an Inductive reasoning mental model increased with grade level. Note that we had only two grade-three participants.**

## 7.1 Tension 1: Overlap and Gaps Between Children's Data, Computational, and AI Literacies

Understanding AI concepts is not a singular skill but rather an intersection of multiple literacies, including data and computational literacy. Each of these literacies plays a distinct yet interconnected role in shaping how children perceive and engage with AI technologies. One critical challenge is that children often struggle to integrate these literacies when reasoning about AI. Data literacy, which refers to the ability to understand, interpret, and critically engage with data, is foundational for grasping how machine learning—a core component of AI—operates [5, 11, 15]. However, our findings suggest that children who conceptualized AI reasoning as inherent and deductive did not demonstrate an understanding of the role of data in AI learning. As a result, children tended to describe AI's reasoning as stemming from its ability to retrieve vast amounts of information from the internet rather than as a system engaging in data-driven learning and pattern recognition. This aligns with prior research [57, 67, 77] that found children frequently conceptualize AI as an "omniscient database."

A similar gap emerged between computational literacy and AI literacy. Computational literacy involves computational thinking and using and understanding code to explore and communicate ideas [22]. Many children who viewed AI as reasoning deductively mistakenly believed that AI systems function solely through explicit, predefined instructions, similar to traditional algorithms that execute fixed sets of steps. While these children recognized that AI is programmed by humans [67], they did not acknowledge that AI models also learn from data to generate outputs [57]. This misconception likely stems from their familiarity with rule-based programming, where systems operate strictly within the logic designed by human programmers. These findings highlight a critical tension: while children may develop an understanding of data and computational literacies in isolation, they often they often struggle to integrate these literacies when reasoning about AI. This suggests a need for educational interventions that explicitly bridge the connections between these domains, helping children build a more comprehensive understanding of AI as both a rule-based and data-driven system.

To bridge this gap, several scholars have proposed extending Brennan and Resnick's CT framework [8] to include AI-specific concepts such as classification, prediction (AI infers likely outcomes), and generation (AI synthesizes new content based on learned patterns) [8, 62, 86]. These CT concepts, when coupled with CT practices such as training, validating, and testing models, can help children distinguish between pre-programmed algorithms and data-driven decision-making systems [8, 62, 86]. We suggest that AI reasoning belongs alongside these now well-established concepts of AI literacy and that foundational concepts drawn from both computational literacy and data literacy can help children develop more robust mental models (see Figure 8).

## 7.2 Tension 2: Generalizing AI Reasoning Across Contexts

Children are faced with the challenge of figuring out what AI means in a world where AI models are being embedded into various applications, from recommender systems to social media to AI tutors. This diversity means that children interact with AI through multiple modalities and are no longer limited to engaging with AI via embodied models such as voice assistants or smart toys. Instead, they also encounter AI in search engines, chatbots, game environments, and adaptive learning platforms, where AI functions in less visible but equally impactful ways. Each of these AI systems has a different way of making decisions. This creates a fundamental cognitive challenge: *children's understanding of AI is shaped by fragmented and often contradictory experiences.* They encounter AI across different contexts, but there is no clear or consistent pattern that helps them generalize how AI reasons. As a result, they may apply incorrect mental models in certain situations, leading to misconceptions or overgeneralizations about how AI makes decisions [27, 67, 98].

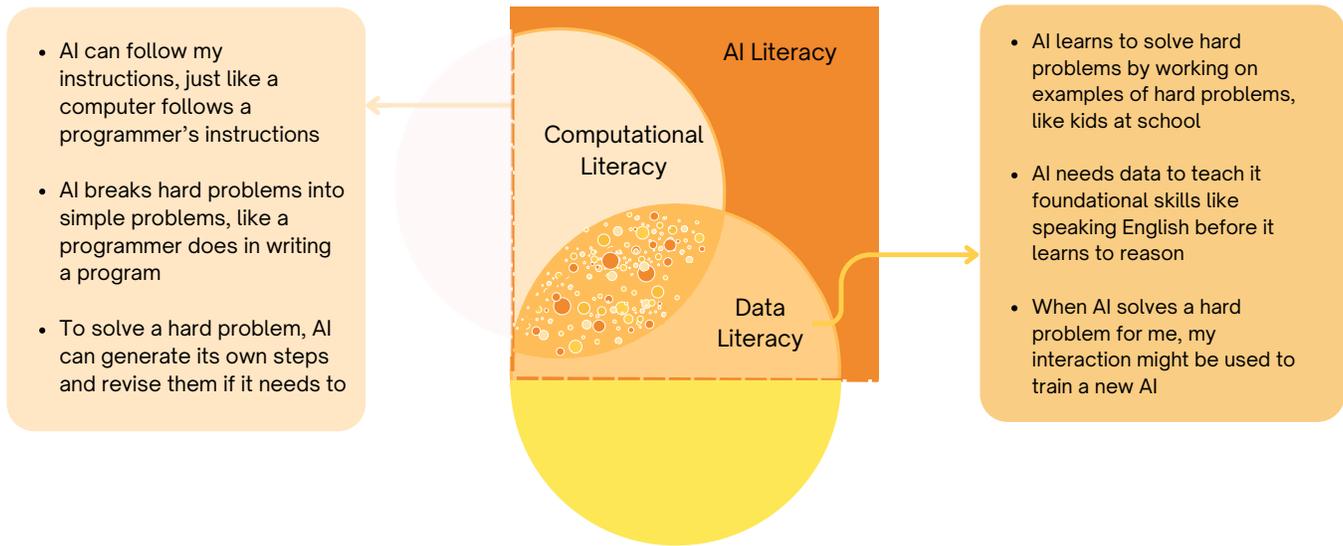

**Figure 8: This figure includes three observations about AI reasoning that would be supported by a background in computational literacy, as well as three that would be supported by a background in data literacy. Our work suggests the potential benefits of a more explicit bridge between these literacies to undergird children's understanding of AI and AI reasoning.**

This raises an important question: *How can we redesign children's digital experiences with AI to scaffold their learning and help them develop a more accurate understanding of AI reasoning?* One promising approach is to integrate AI explainability into these interactions, addressing common misconceptions about how AI makes decisions [41, 74]. This could be scaffolded by helping children first understand broad patterns in the AI system's decision-making. For example, children might first learn that a large language model predicts words based on patterns in large datasets. Once they grasp this general behavior, they can engage with local explainability [41, 53, 74], which focuses on why the AI model made a specific decision in a given case, such as why it misclassified a particular word in a sentence.

### 7.3 Tension 3: Balancing AI Literacy with the Pace of Technological Change

One of the core tensions emerging from our study is the sustainability of AI education in the face of rapid technological advancements. Unlike traditional subjects with relatively stable foundational knowledge, AI is evolving at an unprecedented pace, requiring frequent updates to curricula. This raises key concerns: How can AI education remain up-to-date without overwhelming educators and learners? Is it feasible to design AI literacy programs that continuously adapt without causing cognitive or informational overload for students? Our findings suggest that children's mental models are often built on somewhat outdated or oversimplified understandings of AI's capabilities. If AI education remains static, these misconceptions persist. However, if AI curricula are updated too frequently or introduce too much complexity at once, students (and educators) may struggle to keep pace, leading to frustration, disengagement, or misrepresentation of new information. The tension, therefore, lies in balancing the need for continuously updated AI education with the cognitive and logistical limits of learners and educational systems.

One approach to managing this tension is modular AI education, where lessons are structured in a way that allows for iterative updates without requiring constant full-scale curriculum overhauls. Rather than presenting AI education as a fixed syllabus, educators could adopt an evolving model where fundamental concepts remain consistent, but emerging AI developments are introduced gradually through supplementary modules [21]. This would prevent both educator fatigue (from needing to frequently redesign curricula) and student overload (from being bombarded with constant updates). Additionally, AI literacy could leverage interactive tools that allow students to explore AI's development over time. One approach could be introducing AI model lineages, where students can trace the progression of AI technologies from early models (*e.g.*, ELIZA and rule-based systems) to next generation LLMs like ChatGPT, Gemini, or Claude. By visualizing and interacting with different AI generations, students can develop a historical and conceptual understanding of AI's iterative improvements. This approach situates new AI developments within a broader technological context, allowing students to see the incremental nature of AI progress rather than perceiving AI as an unpredictable and constantly shifting entity.

## 8 Limitations & Future Work

Our study examined how children in grades 3 to 8 conceptualize AI reasoning. This age group has also been widely studied in AI education and human-AI interaction research [24, 26, 28, 56, 57, 83], demonstrating their capacity to engage in meaningful discussions about AI's decision-making processes. At the same time, we acknowledge that this excludes younger children, who may have different mental models of AI reasoning, and high school students, whose understanding may be more advanced. Future work could build on our findings to explore how AI reasoning is conceptualized across the full K-12 spectrum. Another limitation of our study is its geographic scope. While we included children from diverse backgrounds, all participants were from a single region in a large US city. Given that cultural factors may shape how children perceive AI [17], future work could examine whether our findings hold across different cultural contexts.

One misconception we were expecting in our study but was notably sparse was children's reference to robots. Prior studies have highlighted robots as central to children's understanding of AI, often portraying them as symbolic of AI's autonomy and problem-solving capabilities. In our study, while children did describe AI as inherently intelligent, robots were mentioned by only three participants when explaining their reasoning abilities. Additionally, while prior research has found that children often anthropomorphize AI, across all three reasoning models, children in our study framed AI as constrained by its inability to engage with emotions or human experiences. One possible explanation may be that their exposure to ARC puzzles, prior to giving their input, primed the children to think about AI more abstractly. Another possible explanation is the increased visibility of non-embodied AI models, such as generative AI, could be broadening how children conceptualize AI beyond its traditional association with robots. Future research could explore whether increasing exposure to generative AI is reshaping children's mental models of AI.

## 9 Conclusion

Drawing on a common benchmark for assessing AI reasoning capabilities, ARC Puzzles, our work offers an account of children's mental models of AI *reasoning*, a rapidly advancing area of AI research now building on advances in generative AI. Our research indicates that, despite evidence of tensions related to the pace of technological change, problems parsing the numerous new forms of AI, and gaps in children's technological literacies, there are also opportunities for our approaches to children's computational and data literacies to continue evolving to support strong mental models of reasoning technologies, which are likely to have a significant impact on children's lives.

## Acknowledgments

This material is based upon work supported under the AI Research Institutes program by the National Science Foundation and the Institute of Education Sciences, U.S. Department of Education, through Award #DRL-2229873 - AI Institute for Transforming Education for Children with Speech and Language Processing Challenges (or National AI Institute for Exceptional Education). Any opinions, findings, and conclusions or recommendations expressed in this material are those of the author(s) and do not necessarily reflect the views of the National Science Foundation, the Institute of Education Sciences, or the U.S. Department of Education. This work was also partially funded by the Jacob's Foundation CERES Network.

## 10 Selection and Participation of Children

### 10.1 Participation in the Co-Design Study

Children who participated in our co-design study were involved in an intergenerational co-design group at our university. Prior to participation, parental consent and child assent were obtained, and assent forms were written using age-appropriate language. Consent and assent forms were approved by our IRB. Parents and children were fully briefed on the study's objectives, potential risks, and confidentiality protocols. They were also informed that participation was entirely voluntary, and children had the freedom to withdraw at any point. All adult facilitators completed institutional training on ethics and child safety. Children's data was anonymized and stored securely.

### 10.2 Participation in the Field Study

The field study took place during UW Discovery Days, a public STEM outreach event at our university that attracted a large and diverse group of children from the local community. Before participating, each child was given a brief introduction to the study, and their chaperones (teachers, parents, or guardians) were informed about the research goals. While our university's IRB determined that the study met the criteria for exemption due to its educational nature, we nonetheless obtained written assent from all children and verbal consent from their chaperones to ensure informed participation. All participants were explained that participation was voluntary and that they could withdraw at any time. Children's data was anonymized and stored securely.